\documentclass[10pt,letterpaper]{article}

\usepackage{cvpr}
\usepackage{microtype}
\usepackage{xcolor}
\usepackage{tcolorbox}
\usepackage{enumitem}
\usepackage{tcolorbox}
\usepackage[table]{xcolor}
\usepackage{bm}
\usepackage{url}
\usepackage{etoc}
\usepackage{algorithm}       
\usepackage{natbib}
\usepackage{bbding}
\usepackage{makecell}
\usepackage{arydshln}
\usepackage{pifont}
\usepackage{algpseudocode}
\usepackage{booktabs}   
\usepackage{graphicx}   
\usepackage{multirow}
\usepackage{pifont}
\usepackage{subcaption}
\usepackage{wrapfig}
\usepackage{amsthm}
\usepackage{enumitem}

\usepackage{marvosym}
\usepackage{multirow} 
\usepackage{makecell}
\usepackage{tabularx}
\usepackage{newfloat}
\usepackage{listings}
\usepackage{arydshln}
\usepackage{pifont}
\usepackage{algorithm}
\usepackage{algpseudocode}

\definecolor{GreenCheck}{RGB}{0, 102, 51}

\definecolor{customblue}{HTML}{2C70C9}
\definecolor{customred}{HTML}{E74A7A}
\definecolor{imgtokenblue}{HTML}{14B6F1}
\definecolor{rebeccapurple}{HTML}{663399}
\definecolor{cvprblue}{rgb}{0.21,0.49,0.74}

\makeatletter
\renewcommand{\fnum@figure}{\figurename~\textcolor{red}{\thefigure}}
\renewcommand{\fnum@table}{\tablename~\textcolor{red}{\thetable}}

\makeatother

\usepackage[pagebackref,breaklinks,colorlinks,allcolors=cvprblue]{hyperref}

\let\origrmdefault\rmdefault
\let\origsfdefault\sfdefault

\usepackage{libertine}            

\renewcommand{\rmdefault}{\origrmdefault}
\renewcommand{\sfdefault}{\origsfdefault}
\normalfont

\newcommand{\PaperTitle}{Think Before You Drive: World Model-Inspired \\Multimodal Grounding for Autonomous Vehicles}

\newcommand{\PaperAuthors}{\textbf{Haicheng Liao}$^1$\textsuperscript{\href{mailto:yc27979@um.edu.mo}{\Cancer}}, \textbf{Huanming Shen}$^2$\textsuperscript{\href{mailto:huanmingshen@outlook.com}{\Cancer}}, \textbf{Bonan Wang}$^1$, \textbf{Yongkang Li}$^3$, \\ \textbf{Yihong Tang}$^4$, \textbf{Chengyue Wang}$^1$, \textbf{Dingyi Zhuang}$^5$, \textbf{Kehua Chen}$^6$, \\
\textbf{Hai Yang}$^7$, \textbf{Chengzhong Xu}$^1$\textsuperscript{\href{mailto:czxu@um.edu.mo}{\Letter}}, \textbf{Zhenning Li}$^1$\textsuperscript{\href{mailto:zhenningli@um.edu.mo}{\Letter}}}

\newcommand{\PaperAffiliations}{$^1$University of Macau,\quad $^2$UESTC,\quad $^3$Purdue University,\quad$^4$McGill University,\quad$^5$Massachusetts Institute of Technology,\quad $^6$University of Washington,\quad$^7$The Hong Kong University of Science and Technology}

\newcommand{\AuthorNotes}{\textsuperscript{\Cancer}Equal Contribution, \quad \textsuperscript{\Letter}Corresponding Authors}

\newcommand{\PaperDate}{\textbf{Date}: November 25, 2025}

\newcommand{\PaperAbstract}{%
Interpreting natural-language commands to localize target objects is critical for autonomous driving (AD). Existing visual grounding (VG) methods for autonomous vehicles (AVs) typically struggle with ambiguous, context-dependent instructions, as they lack reasoning over 3D spatial relations and anticipated scene evolution. Grounded in the principles of world models, we propose ThinkDeeper, a framework that reasons about future spatial states before making grounding decisions. At its core is a Spatial-Aware World Model (SA-WM) that learns to reason ahead by distilling the current scene into a command-aware latent state and rolling out a sequence of future latent states, providing forward-looking cues for disambiguation. Complementing this, a hypergraph-guided decoder then hierarchically fuses these states with the multimodal input, capturing higher-order spatial dependencies for robust localization. In addition, we present DrivePilot, a multi-source VG dataset in AD, featuring semantic annotations generated by a Retrieval-Augmented Generation (RAG) and Chain-of-Thought (CoT)-prompted LLM pipeline. Extensive evaluations on six benchmarks, ThinkDeeper ranks~\#1 on the Talk2Car leaderboard and surpasses state-of-the-art baselines on DrivePilot, MoCAD, and RefCOCO/+/g benchmarks. Notably, it shows strong robustness and efficiency in challenging scenes (long-text, multi-agent, ambiguity) and retains superior performance even when trained on 50\% of the data.
}

\definecolor{metabg}{HTML}{F0F5F7} 
\definecolor{metafg}{HTML}{1C2B33} 
\newcommand{\TitleBox}{

  \begingroup
  \renewcommand{\rmdefault}{LinuxLibertineT-TLF}
  \renewcommand{\sfdefault}{LinuxBiolinumT-TLF}
  \normalfont
  
  \begin{tcolorbox}[
    colback=metabg,
    colframe=white,
    boxrule=0pt,
    arc=10pt,
    left=0.8cm,
    right=0.8cm,
    top=0.8cm,
    bottom=0.8cm
  ]
    {\LARGE\bfseries\sffamily \PaperTitle\par}
    \vspace{1.2em}

    {\large\sffamily \PaperAuthors\par}
    \vspace{0.6em}

    {\normalsize\sffamily \PaperAffiliations\par}
    \vspace{0.5em}

    {\normalsize\sffamily \AuthorNotes\par}
    \vspace{1.0em}

    {\large\bfseries\sffamily Abstract\par}
    \vspace{0.4em}

    {\color{metafg}\PaperAbstract\par}
    \vspace{0.8em}
    
    {\small\sffamily \PaperDate\par}
    
  \end{tcolorbox}
  
  \endgroup
  \vspace{1.5em}
}

\begin{document}
\twocolumn[
\begin{@twocolumnfalse}
\TitleBox
\end{@twocolumnfalse}
]

\section{Introduction}\label{sec:intro}
\begin{figure}[t]
  \centering
\includegraphics[width=\linewidth]{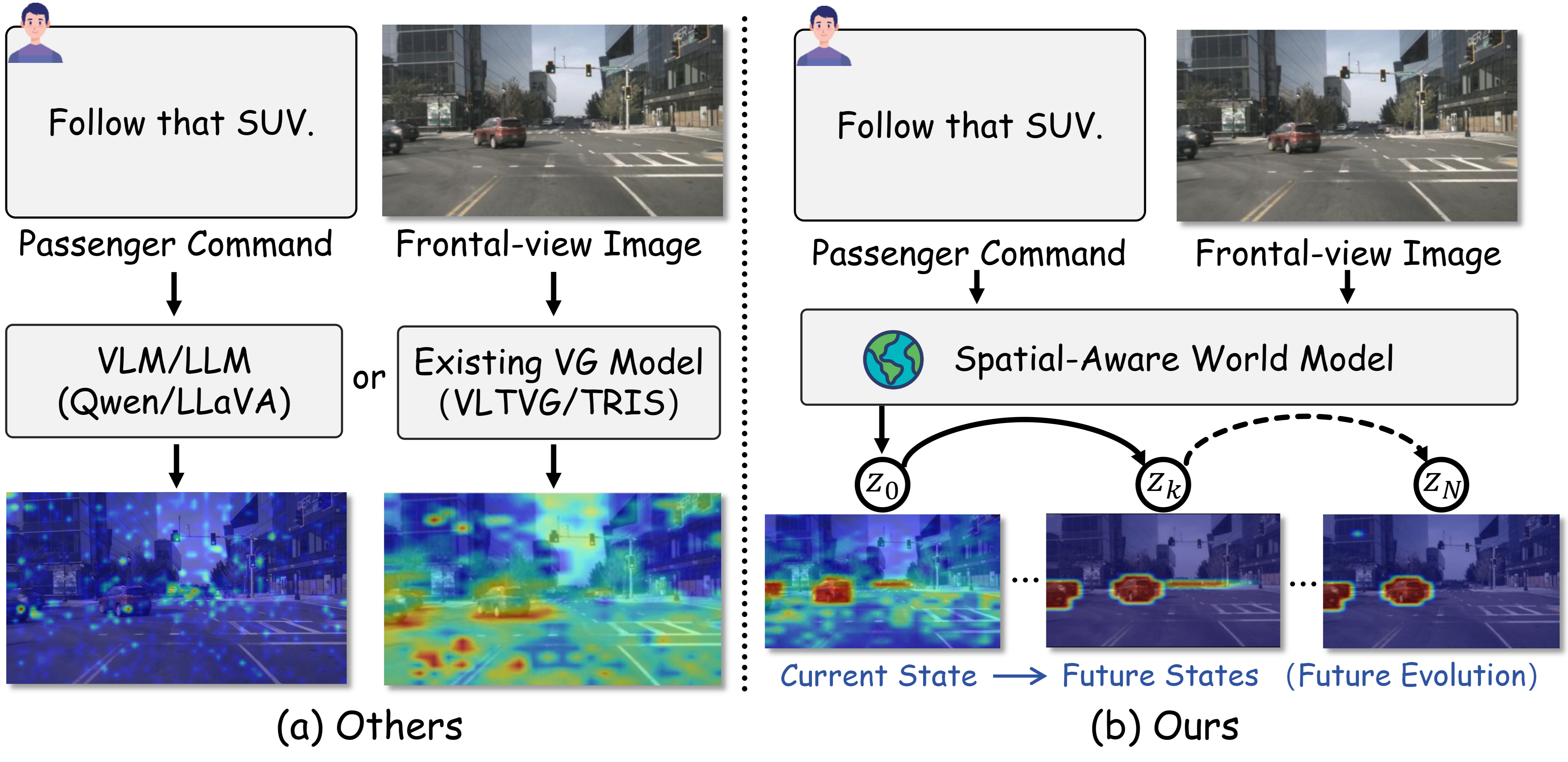}
  \caption{Comparison of visual grounding performance between existing VLMs and our model in real-world driving scenarios. (a) General-purpose VLMs (Qwen~\cite{wang2024qwen2}/LLaVA~\cite{liu2024llavanext}) and existing VG models~\cite{yang2022improving,liu2023referring} typically fail to robustly localize targets under challenges such as motion blur, ambiguous language, and multi-agent traffic contexts. (b) In contrast, our Spatial-Aware World Model first distills the scene and command into a current latent state ($z_0$), and then reasons ahead by rolling out a sequence of future latent states ($z_k \rightarrow z_N$), obtaining a forward-looking perspective that enables more reliable and spatially coherent grounding.}
  \label{toutu}
\end{figure}

In recent years, the pursuit of fully autonomous vehicles (AVs) has captivated both industry and academia \cite{wang2025wake}. Despite remarkable technological advances, widespread public acceptance remains elusive, primarily due to concerns over the reliability of human-machine interaction and apprehensions about losing control. These challenges are magnified in complex scenes where vehicles must make split-second decisions, highlighting the need for enhanced communication between humans and machines. Visual grounding (VG) in AVs, where vehicles interpret and act on natural language commands, emerges as a pivotal innovation, empowering passengers with a direct and intuitive mode of interaction that significantly enriches the driving experience. VG in autonomous driving (AD) requires a deep understanding of both the immediate and forthcoming environment \cite{dai2024simvg}. For instance, when a passenger says ``merge behind the white SUV after the crosswalk'', the AV must jointly reason about spatial relationships and textual intent: the distance to two similar SUVs at different depths, the approach of a cyclist entering the blind spot, and the state of a traffic signal that reduced to a few pixels in its field of view \cite{krugel2022autonomous, faulhaber2019human}. This grounding challenge is further complicated by the inherent variability and ambiguity of natural language commands, which are often highly dependent on the driving context.

Traditional VG methods \cite{zhan2024mono3dvg,geng2025pseudo,jiang2022pseudo} are ill-suited for the demands of real-world AD due to two primary issues. First, they are typically designed for high-resolution, controlled datasets, and struggle with the challenging visual conditions encountered on the road, such as low light and fast-moving scenes. These real-world conditions often obscure critical details, causing the methods to frequently miss vital contextual cues in ambiguous or rapidly changing situations. Second, existing VG models~\cite{liao2024and,10342826,LIAO2024100116,chen2024mpcct,chen2023minigptv2,liu2023grounding} generally lack the 3D spatial awareness required for complex relational reasoning. This lack of spatial intelligence limits their ability to accurately assess object distances or differentiate between immediate hazards and distant background elements. This shortcoming is evident in context-dependent commands like ``Avoid the cyclist ahead'', where the model must determine which cyclist is in immediate proximity and requires action, rather than simply identifying any cyclist in the scene~\cite{liao2025cot}. 

In parallel, recent studies \cite{wang2024qwen2,liao2025cot,ma2025position} have integrated Large Language Models (LLMs) and Vision-Language Models (VLMs) to improve semantic understanding. While these models excel at resolving ambiguities via deep reasoning, they introduce significant drawbacks: massive data requirements, prohibitive computational costs, and high inference latency, impeding real-time deployment in AD. These limitations provoke us to ask an important question: \textit{How can we design a VG model for AD that is spatially aware, robust to ambiguity, and efficient enough for real-time operation?}

\begin{figure}[t]
  \centering
  \includegraphics[width=1.0\linewidth]{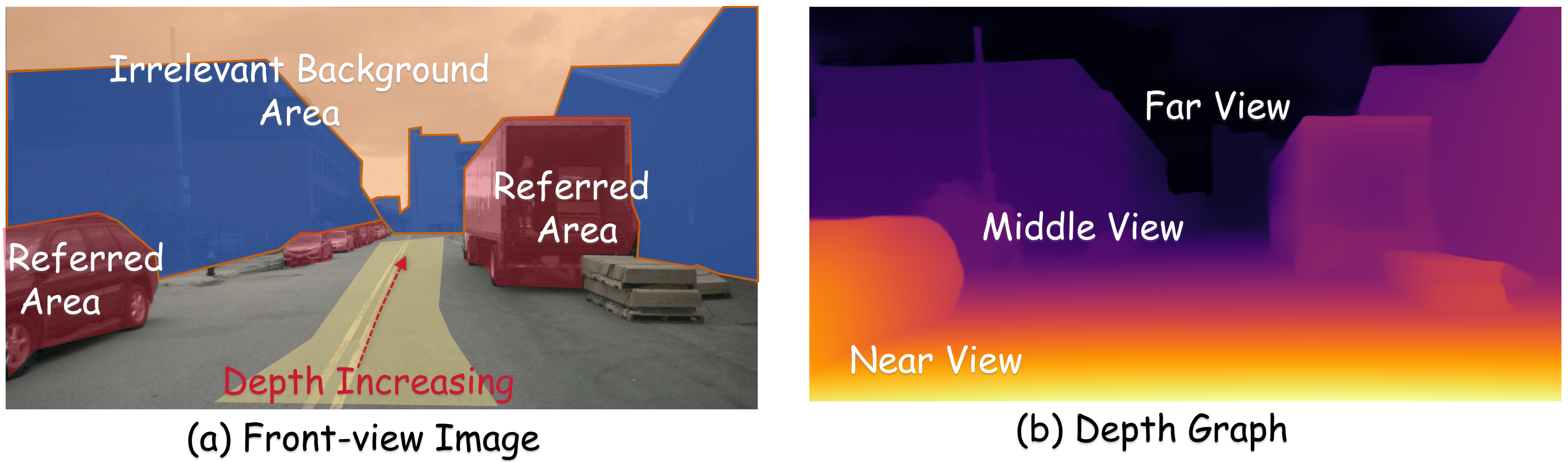}
  \caption{Illustration of depth-based spatial priors. (a) Real-world scenes have a clear 3D structure, with referred areas (nearby vehicles) at different distances than the irrelevant background (distant sky, mid-range buildings). (b) The depth map indicates depth semantics across near, middle, and far views, enabling the SA-WM to enhance its spatial awareness and filter implausible regions.}
  \label{depth_map}
\end{figure}

To alleviate this problem, we introduce a world model-based framework that empowers AVs to ``\textbf{think deeper}" by reasoning about how the future scene is likely to evolve, effectively bridging the gap between rigid algorithmic processing and human-like contextual reasoning. Specifically, we introduce a Spatial-Aware World Model (SA-WM) to predict future states from the current scene and command, providing a forward-looking perspective for the grounding decision. As shown in Figure \ref{toutu}, it first constructs a command-aware current latent state that filters out irrelevant elements (e.g., roadside buildings) and then iteratively rolls out future latent states to foresee critical cues essential for decision-making. By reasoning over these prospective states before decoding, the model attains more stable decisions that improve generalization and safety in AD grounding. To enhance the model's spatial awareness, we integrate monocular depth to provide 3D localization for the vision-only pipeline. As illustrated in Figure \ref{depth_map}, this depth signal enables the SA-WM to prioritize entities by proximity and relevance, mimicking human spatial perception. Complementing this, we introduce a cross-modal hypergraph decoder that captures higher-order relations between textual phrases and spatial regions, effectively fusing the predicted future latent states from the SA-WM to achieve robust grounding.
Overall, the main contributions of this study can be summarized as follows:
\begin{itemize}
\item We introduce \textbf{ThinkDeeper}, the first world model-based framework for visual grounding in AD. ThinkDeeper predicts future latent states from the current scene, provides a forward-looking perspective to guide AVs to reason about the environment's future evolution before grounding.

\item We present \textbf{DrivePilot}, a multi-source dataset featuring semantic annotations generated via Retrieval-Augmented Generation (RAG) and Chain-of-Thought (CoT) prompting with LLMs. It provides a robust benchmark of dynamic, real-world driving scenes to advance VG research.
 
\item ThinkDeeper sets the state-of-the-art (SOTA) on the DrivePilot, MoCAD, and Talk2Car benchmarks, generalizing to excel on the RefCOCO/+/g datasets. Notably, it remains robust and efficient in scenes, outperforming most top baselines even with 50\% and 75\% training data.

\end{itemize}

 \begin{figure*}[t]
  \centering
  \includegraphics[width=0.95\linewidth]{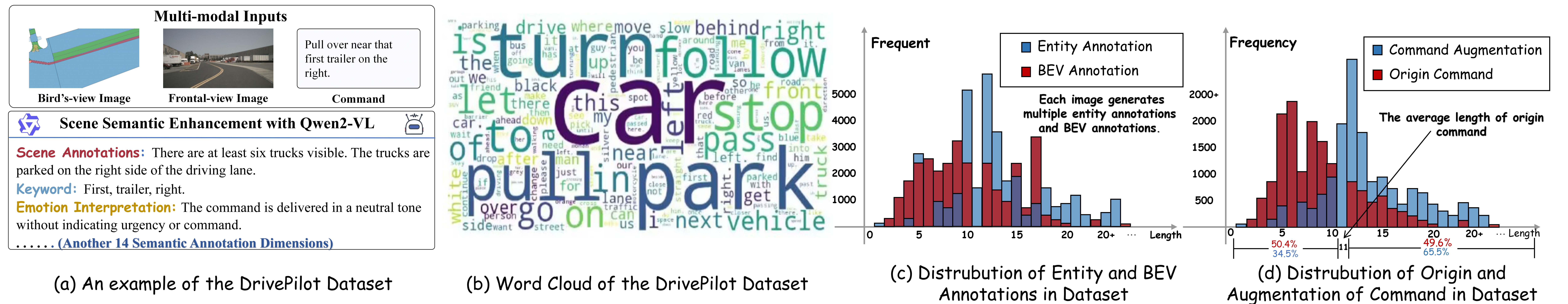}
  \vspace{-5pt}
  \caption{Overview of the proposed DrivePilot. (a) An example of the multi-source representation for a real-world scene, including RAG- and CoT-enhanced annotations (14 semantic dimensions) from Qwen2-VL. (b) Statistics of DrivePilot on the word cloud of the command distributions. (c) Distribution of entity and BEV annotations per image. (d) Length distribution of original and augmented commands.}
    \label{fig3}
\end{figure*}

\section{Related Work}
\label{sec:formatting}
\textbf{Visual Grounding in Autonomous Driving.}
Visual grounding has emerged as a critical component for enhancing human-machine interaction in AD. Early VG methods in this field can be categorized as one-stage or two-stage. VG methods in AD were initially divided into one-stage and two-stage methods. One-stage models \cite{liao2020real,yang2020improving,yang2022improving} are known for their efficiency, processing images and commands in a unified architecture. However, they may falter in scenarios with densely populated or overlapping regions. In contrast, two-stage models \cite{tan2019lxmert,chen2021ref} first generate a set of region proposals using a pre-trained detector and then match these regions against the text. While often more precise, their performance is fundamentally capped by the quality of the initial region proposals. Recent work has employed VLMs like Qwen2-VL \cite{wang2024qwen2}, and MiniGPT-v2 \cite{chen2023minigptv2} for their strong cross-modal semantic reasoning. However, their prohibitive computational overhead and high inference latency are incompatible with the real-time constraints of autonomous driving systems. Consequently, this progression highlights a critical and unresolved gap: the need for developing a framework that is both semantically robust for complex driving scenes and efficient enough for real-time, on-board computation.\\
\textbf{World Model in Autonomous Driving.} With the rapid progress of LLMs and modern generative models, world models have emerged as a new paradigm for model-based prediction and planning \cite{ding2025understanding}. World models learn a compact latent representation of the external scenes, enabling an agent to roll out imagined future states and evaluate candidate actions before making a decision \cite{ha2018recurrent}. Current applications of world models in AD fall into three main categories \cite{guan2024world}. First, for end-to-end driving \cite{chen2025drivinggpt,gao2024enhance,zheng2025world4drive}, models
like LAW \cite{li2024enhancing} and Drive-WM \cite{wang2024driving} leverage a world model to predict future scene states, providing a robust, forward-looking context for downstream decision-making.
Second, for scenario simulation \cite{guan2025world,wang2024drivedreamer,ni2025recondreamer}, models such as DriveDreamer-2 \cite{zhao2025drivedreamer} and Vista \cite{gao2024vista} utilize diffusion models or MLLMs to generate physically plausible video sequences for closed-loop simulation. Third, in representation learning \cite{lu2025infinicube,zheng2024occworld,zuo2025gaussianworld}, works like DriveDreamer4D \cite{zhao2025drivedreamer4d} and FSDrive \cite{zeng2025futuresightdrive} use world models to couple geometry, appearance, and agent dynamics for structured 3D/4D priors.
However, how world models can facilitate VG tasks for AD remains unexplored. To our knowledge, we are the first to explore the effective application of the world model to VG tasks for autonomous vehicles, presenting preliminary studies in this emerging field.

\begin{figure*}[t]
  \centering
  \includegraphics[width=0.96\linewidth]{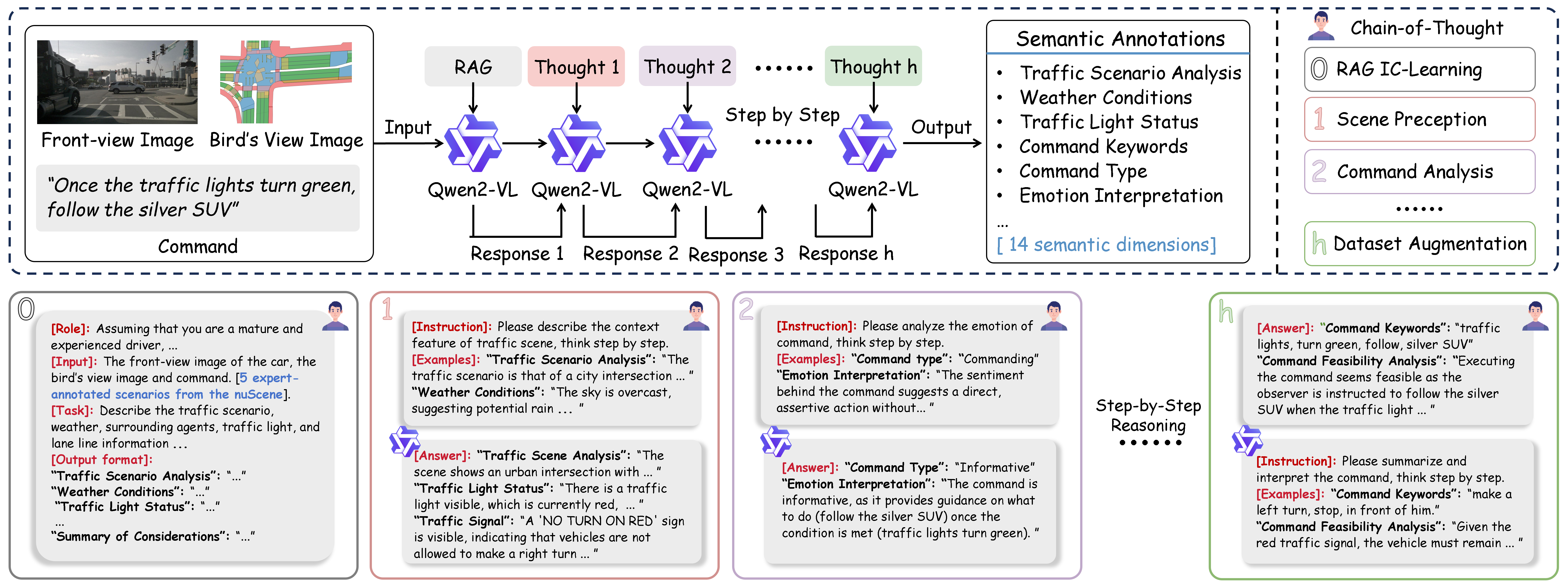}
  \vspace{-3pt}
  \caption{Illustration of the CoT prompting used in DrivePilot to generate semantic annotations for a given traffic scene. This step-by-step process involves dialogues where each ``thought'' guides the Qwen2-VL to understand different aspects of the scene or given command.}
    \label{cot}
  \end{figure*}

\section{DrivePilot Dataset}
We introduce DrivePilot, a large-scale VG dataset for AD research based on nuScenes \cite{caesar2020nuscenes}. It provides a rich collection of real-world scenes from urban (Singapore and Boston), capturing a broad spectrum of driving conditions, weather patterns, and diurnal cycles. DrivePilot is pioneering in leveraging the capabilities of Qwen2-VL \cite{wang2024qwen2} to generate semantic annotations. To enhance the LLM's scene comprehension and minimize hallucinations, we employ RAG \cite{yuan2024rag} to dynamically retrieve relevant external information. Furthermore, we use a zero-shot CoT prompting strategy to guide the LLM in generating step-by-step semantic annotations. The generation process is outlined in three steps:

\noindent\textbf{Step-1: In-Context RAG Annotation.}
We first construct a knowledge base from 1,200 curated nuScenes samples, detailing agent trajectories, agent types, and road conditions. For each new scene to be annotated, we retrieve the top-$k$ relevant scenarios from this knowledge base via cosine similarity. These retrieved examples serve as in-context cues (e.g., historical vehicle behavior in similar weather, pedestrian patterns) to guide the Qwen2-VL in generating context-aware, structured scene annotations for our dataset.

\noindent\textbf{Step-2: CoT Prompting Annotation Generation.} 
This step unfolds as a progressive dialogue, with each step directing the LLM to focus on distinct facets of the scene. As depicted in Figure \ref{cot}, this CoT process involves a structured dialogue where each ``thought'' guides the Qwen2-VL's reasoning. The first thought focuses on understanding the overall scene, identifying key objects, and indicating their spatial relationship. The second thought analyzes command keywords and semantic intent. Subsequent thoughts prompt the Qwen2-VL to progressively consider factors such as road conditions, traffic density, notable events, and the potential behaviors of each agent. After $h$ reasoning iterations (where $h$ varies by sample), the insights from this chain of thought are synthesized into a cohesive semantic annotation, which is then formatted for manual verification.

\noindent\textbf{Step-3: Manual Cross-check Validation.}  
All annotations generated from Qwen2-VL undergo manual verification by a panel of 13 domain experts, including AV safety engineers, certified driving instructors, and postgraduate researchers. Each sample is assessed for consistency with ground-truth sensor data and adherence to the local traffic laws. In addition, any discrepancies, such as misaligned object references, trigger a re-annotation process to ensure all labels meet real-world legal and operational standards. 

As shown in Figure \ref{fig3}, DrivePilot is a comprehensive benchmark for a range of AV tasks, including command understanding, object detection, and visual grounding. It incorporates 14 semantic dimensions for contextual richness, such as weather, traffic light status, and emotional context, to ensure semantic richness. Moreover, BEV images are formatted to consistently depict the target vehicle oriented to the right, with a resolution of 1200×800, while front-view images are calibrated to 1600×900. Each dataset entry is a comprehensive sample, comprising a natural language command, paired front-view and BEV images, LLM-generated scene annotations, and the precise target object location. Each entry includes a natural language command (averaging 14.72 words), LLM-generated scene annotations, and precise target locations. By providing these diverse contexts and challenging language-vision cases, DrivePilot serves as a new standard to accelerate progress in AD grounding.

\section{Methodology}\label{Methodologies}
The goal of this study is to create a model capable of interpreting a frontal-view image, $I$, and a natural language command $C$. The model's task is to pinpoint the specific area within the image $I$ that corresponds to the destination or target object described in the command $C$. This requires the model to execute advanced cross-modal reasoning by seamlessly blending visual cues from $I$ and interpretative insights from command $C$. In a nutshell, the model aims to determine the exact destination or object the AD is instructed to approach or identify based on the given command.

\begin{figure*}[t]
  \centering
  \includegraphics[width=0.95\linewidth]{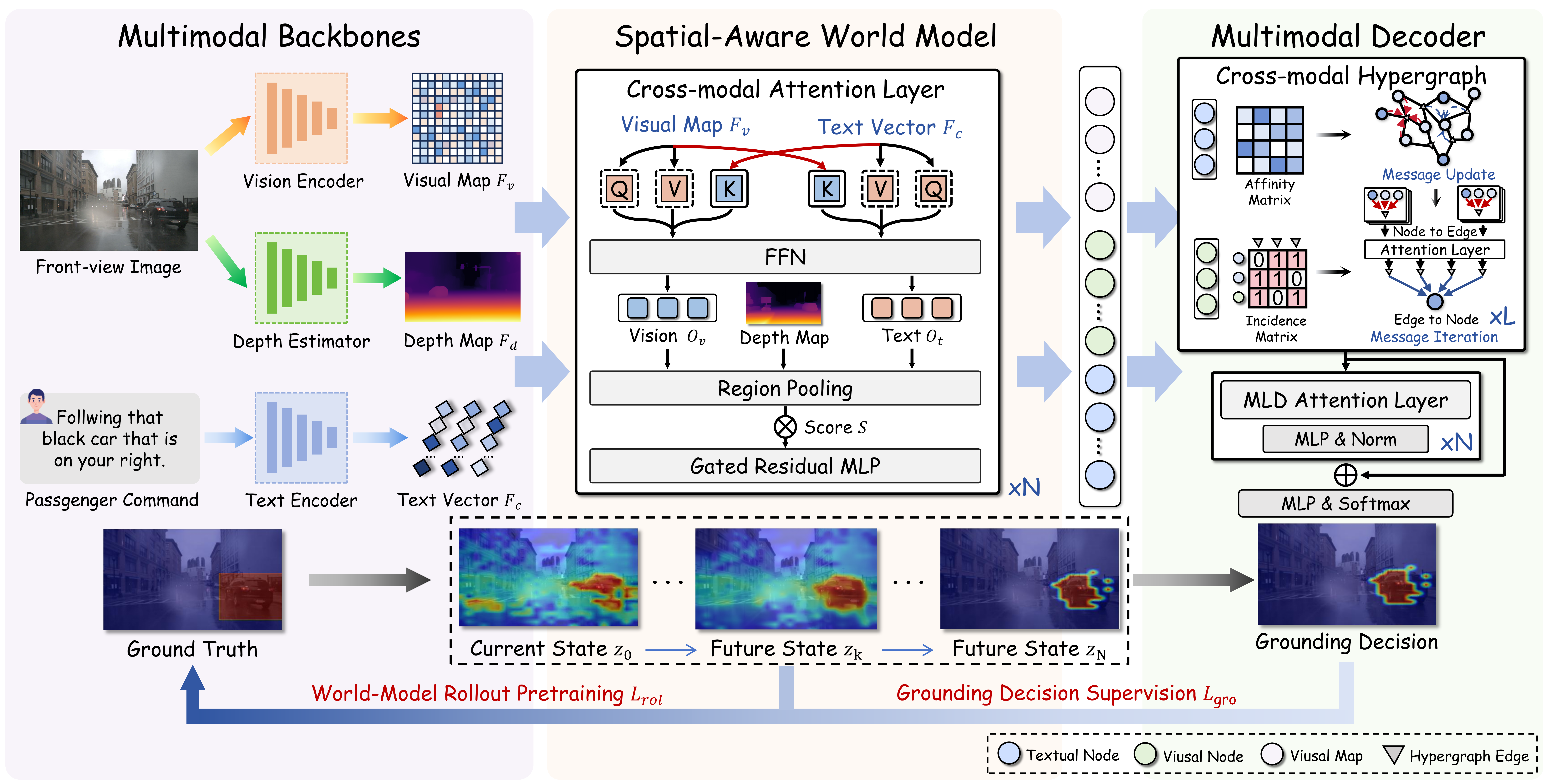}
  \caption{Overview of ThinkDeeper. Multimodal Backbones first encode the image $I$ and command $C$ into visual $F_{v}$, textual $F_{t}$, and depth $F_{d}$  features. Next, the SA-WM takes these features and reasons ahead by rolling out a sequence of future latent states ($z_0 \rightarrow z_N$). Finally, the decoder applies a cross-modal hypergraph network to fuse these predicted states and produce the final, robust grounding decision $\bm{Y}$.}
\label{fig:4}
\end{figure*}

\subsection{Overall Pipeline}
Figure \ref{fig:4} shows the pipeline of ThinkDeeper,  which departs from traditional ranking-based visual localization and comprises three components: (i) Multimodal Backbones, (ii) Spatial-Aware World Model, and (iii) Multimodal Decoder. Initially, the backbones extract the frontal-view image $I$ and command $C$ into rich vectors. Then, the SA-WM leverages these vectors to build a compact scene latent state that preserves salient visual cues while filtering background clutter, and iteratively rolls out plausible future latent states $X_v$ to inform downstream grounding. Finally, the Multimodal Decoder fuses the predicted states with multimodal features and reasons across modalities to localize the object that best matches the natural command given by the passengers.

\subsection{Multimodal Backbones}

\noindent\textbf{Vision Encoder.}  
We employ a stack of Vision Transformers (ViTs) \cite{dosovitskiy2020image} to encode the image $I$ into into a visual feature map $F_{v} \in \mathbb{R}^{D \times H \times W}$, where $D$ is the the channel dimension and $H \times W$ matches the image resolution. In parallel, two specialist networks provide auxiliary cues: CenterNet \cite{zhou2019objects} detects objects and yields an object vector $F_o \in \mathbb{R}^{N \times D}$ for the top-$N$ objects, while ZoeDepth \cite{bhat2023zoedepth} estimates monocular depth, resulting in a depth map $F_d \in \mathbb{R}^{H\times W}$ aligned with the spatial dimensions of the visual feature map $F_{v}$.

\noindent\textbf{Text Encoder.} 
The command \(C\) is tokenized with BERT’s WordPiece tokenizer \cite{devlin2018bert} and encoded by a BERT encoder, yielding text vector \(F_{c}=\{x_1, x_2, \ldots, x_L\}\in \mathbb{R}^{L \times Q}\), where $L$ denotes the token length and $Q$ represents the hidden size. 

\subsection{Spatial-Aware World Model}
This component is responsible for (i) distilling the current scene into a command-aware latent state and (ii) rolling this latent forward to envisage future latent states that guide grounding. Concretely, the SA-WM operates in two phases: (1) Current State Construction and (2) Future States Rollout.

\noindent\textbf{Current State Construction.} As shown in Figure \ref{fig:4}, the first phase constructs a compact latent state $z_0$ that represents the current scene. Given the visual feature map $F_v$, and the depth map $F_d$, a set of cross-modal attention layer projects them into a unified semantic space, producing vision-text vectors $O_{\textit{t}} \in \mathbb{R}^{M \times L}$ and $O_{\textit{v}} \in \mathbb{R}^{D \times H \times W}$. Formally,

\vspace{-6pt}
{\begin{small}
\begin{equation}
\mathbf{A}_t = {\phi_{\text{Softmax}}} \left( (F_{v} \mathbf{W}_q^{vis}) \otimes (F_{c} \mathbf{W}_k^{t})^T / \sqrt{D} \right)
\end{equation}
\end{small}}
\vspace{-2pt}
{\begin{small}
\begin{equation}
\mathbf{A}_v = {\phi_{\text{Softmax}}} \left( (F_{c}\mathbf{W}_q^{tex}) \otimes (F_{v} \mathbf{W}_k^{vis})^T / \sqrt{D} \right)
\end{equation}
\end{small}}
\vspace{-2pt}
{\begin{small}
\begin{equation}
O_{\textit{t}} = \mathbf{A}_{t}^T \otimes (F_{v}\mathbf{W}_{v}^{vis}), \quad O_{\textit{v}} = \mathbf{A}_v^T \otimes (F_{c}\mathbf{W}_{v}^{tex})
\end{equation}
\end{small}}

Here, $\mathbf{A}_v$ and $\mathbf{A}_t$ are the affinity propagation of text-to-visual features and visual-to-text features, respectively. $\otimes$ donates the matrix multiplication, while $\mathbf{W}_{c}^{vis}$ and $\mathbf{W}_{v}^{tex}$ are learnable parameters for $F_v$ and $F_c$, respectively. To distill this rich representation into a compact state $z_t$ and filter out background clutter, we compute a fine-grained saliency score $s^k$ for each visual patch $k$. Mathematically,

\begin{small}
\begin{equation}
s^k = \frac{\sigma^2 \cdot \vec{\mathbf{a}}^TP(k)}{  \exp \left(  \left(1 - \sum_{j} {F_{v}(k, j)^{\mathrm{T}}  O_{\textit{v}}(k, j)}\right)^2 / 2\mu \right)}
\label{eq:scores}
\end{equation}
\end{small}
\vspace{-4pt}

Here, $P(k)$ is a depth-derived prior from \(F_d\) that biases attention toward physically plausible regions, $\mu$ and $\sigma$ are learnable parameters, and $j$ is the $j$-th channel of dimension $D$.  The saliency map at layer $l$ is denoted $s^{(l)}= \{s^k\}_{k=1}^{n}$, and all layer-wise maps are collected as score $S = \{s^{(1)}, \dots, s^{(L)}\}$. Regions with low text–visual affinity or inconsistent depth geometry are assigned low scores and suppressed. Subsequently, we apply Region Pooling \cite{girshick2015fast} over the candidate regions on score $S$ to obtain an aggregated map $\tilde{S}$, which is broadcast and used to gate the object vector $F_o$, generating the current latent state: $z_0 = \phi_{\text{MLPs}} \left(F_{o} \odot \tilde{S} +F_{o}\right)$, where $\odot$ is the Hadamard product and $\phi_{\text{MLPs}}$ is the linear projection. The resulting latent highlights command-relevant structure (objects, geometry, intent cues), while suppressing background clutter.

\noindent\textbf{Future States Rollout.} 
Grounding in dynamic scenes requires not only an accurate understanding of the current scene but also forward-looking reasoning to anticipate future developments. Accordingly, the second phase predicts a sequence of future latent states that think visually about how the scene is likely to evolve under the command. Starting from the latent state $z_0$, a gated residual MLP realizes the recurrent transition $f_\theta$ in latent space: $z_{k+1}=f_\theta\big(\{z_{k}\}_{k=1}^{N-1},\ O_t\big)$,  step-by-step producing the future latent states $Z_{v}= \{z_{1}, z_{2}, \dots, z_{N}\}$. Here, $O_t$ provides language conditioning and encodes geometric constraints. The predictions are in latent space rather than pixels, capturing prospective saliency, geometry-aware attention, and intent cues most informative for the final grounding process. 
The Multimodal Decoder then consumes $Z_{v}$ to localize the target with spatial consistency. 
See \textbf{Appendix~\ref{SA_WM}} for details.
\begin{table*}[t]
  \centering
     \setlength{\tabcolsep}{3.5mm}
\setlength\tabcolsep{6pt}
     \resizebox{1\linewidth}{!}{
    \begin{tabular}{c|c|c|c|c|c|c|c|c|c|c}
      \bottomrule
     \multirow{2}[0]{*}{Model} & \multirow{2}[0]{*}{Backbone} &  \multirow{2}[0]{*}{Talk2Car} & \multicolumn{2}{c|}{MoCAD} & \multicolumn{2}{c|}{DrivePilot} & \multicolumn{3}{c|}{Corner-case Test sets} & Long-text \\
\cline{4-11} &     &   & test  & val   & test  & val   & Visual Const.& Multi-agent & Ambiguous & val \\
    \hline
    AttnGrounder \cite{mittal2020attngrounder} & ResNet-50 & 61.32  & 62.34  & 64.35  & 62.31  & 64.57  & 62.74  & 64.82  & 64.31  & 57.25 \\
    CMSVG \cite{rufus2020cosine} & EfficientNet & 68.61  & 67.66  & 68.47  & 68.87  & 69.93  & 69.39  & 66.77  & 67.83  & 62.21 \\
    TransVG \cite{deng2021transvg} & ResNet-101 & 65.83  & 68.14  & 70.85  & 66.52  & 68.42  & 68.12  & 66.34  & 69.25  & 65.45 \\
    CMRT \cite{luo2020c4av}  & ResNet-152 & 69.11  & 69.42  & 68.83  & 69.54  & 70.37  & 67.12  & 66.20  & 62.23  & 64.25 \\
    MDERT \cite{kamath2021mdetr} & ResNet-101 & 70.52  & 66.74  & 70.23  & 71.35  & 72.15  & 68.35  & 65.37  & 68.38  & 62.72 \\
    VL-BERT \cite{dai2020commands}& ResNet-101 & 70.03 & 71.42  & 70.54  & 71.47  & 72.36  & 70.29  & 70.14  & 69.84  & 66.70 \\
    RSD-LXMERT \cite{chan2022grounding} & ResNet-101 & 72.64  & 72.35  & 71.46  & 73.37  & 74.52  & 70.22    & 71.87  & 63.44  & 65.80 \\
    VLTVG \cite{yang2022improving} & ResNet-101 & 63.33  & 67.14  & 68.26  & 65.37  & 68.49  & 68.51  & 66.22  & 70.24  & 68.80 \\
    UNINEXT \cite{yan2023universal} & ViT & 70.87  & 70.62  & 71.34  & 71.35  & 73.47  & 69.26  & 68.78  & 71.29  & 65.32 \\
    CAVG \cite{LIAO2024100116} & ViT  & 74.62  & 72.44  & 73.25  & 75.52  & 76.48  & 68.39  & 67.36  & 69.45  & 64.36 \\

    \hline

    MiniGPT-v2  \cite{chen2023minigptv2}  & Llama-2 & 61.15 & 60.89 & 61.72 & 62.85 & 63.27 & 59.14 & 58.33 & 55.41 & 56.78 \\

  LLaVA-NeXT (13B) \cite{liu2024llavanext} & LLM & 42.31 & 43.45 & 44.02 & 43.98 & 44.15 & 41.22 & 40.71 & 37.84 & 39.03 \\
    Qwen2.5-VL-7B \cite{wang2024qwen2} & VLM     & 47.31 & 48.20 & 49.10 & 50.06 & 50.84 & 45.12 & 46.37 & 47.05 & 41.92 \\ 
Qwen2.5-VL-72B \cite{wang2024qwen2} & VLM     & 56.17 & 57.10 & 57.85 & 58.92 & 59.74 & 53.43 & 54.25 & 55.17 & 49.83 \\ 
Qwen3-VL-8B~\cite{yang2025qwen3} & VLM     & 56.19 & 57.25 & 58.16 & 59.05 & 59.85 & 53.55 & 54.49 & 55.25 & 50.13 \\

    \hline
    \rowcolor{blue!7} \textbf{ThinkDeeper (50\%)} & ViT+Hyper. & 68.84  & 70.05  & 73.43  & 70.25  & 72.16  & 66.26  & 68.37  & 72.51  & 71.76 \\
    \rowcolor{blue!7} \textbf{ThinkDeeper}  & ViT+Hyper.   & \textbf{76.64}  & \textbf{75.20}  & \textbf{75.76}  & \textbf{77.27}  & \textbf{79.52}  & \textbf{73.53} & \textbf{74.19}  & \textbf{75.26}  & \textbf{74.08} \\
\bottomrule 
    \end{tabular}%
    }
  \vspace{-6pt}
    \caption{\begin{small}Performance comparison of ThinkDeeper (marked in \textcolor{blue!50}{purple}) and SOTA baselines. \textbf{Bold} values are the best performance.\end{small}}\label{tab:t2c}%
  \label{VG}%
\end{table*}%

\subsection{Multimodal Decoder}
The decoder aims to fuse textual intent with the SA-WM’s future latent states $X_{v}$, reason over higher-order relations between phrases and spatial regions, and output a final grounding decision. To this end, we present a cross-modal hypergraph network to model the interplay between visual and textual modalities, guided by the future latent states, enabling multimodal fusion and localization refinement.

\noindent\textbf{Hypergraph Construction.}
We first define the hypergraph \( \mathcal{G} = (\mathcal{V}, \mathcal{E})\), where the node set $\mathcal{V} = Z_v \cup X_t$ comprises $N$ visual nodes $Z_v = \{z_{1}, \ldots, z_{N}\}$ and $L$ textual nodes $X_t = \{x_{1}, \ldots, x_{L}\}$. \(\mathcal{E}\) is the hyperedge set. To couple modalities, we compute an affinity matrix $A_{ij}$ for every vision–text pair $(z_i, x_j)$ in a shared embedding space. We then construct hyperedges based on this matrix. For each visual node $z_i$, we form a corresponding hyperedge $\mathcal{E}_j \in \mathcal{E}$ (where $j$ indexes the hyperedge) by selecting the top-$k$ textual nodes $\{x_k\}$ with the highest affinity $A_{ik}$. Formally, the affinity and the resulting hyperedge feature $\bm{e_j}$ (the average of its constituent text nodes), which can be formally represented as follows:

\vspace{-10pt}
\begin{small}
\begin{equation}
    A_{ij} = \left( \vec{\mathbf{a}}^T \left[ \bm{W}_{v} {z}_{i} \| \bm{W}_t {x}_{j} \right] \right), \,\,
    \bm{e}_j = \frac{1}{|\mathcal{E}_j|} \sum_{k \in \mathcal{E}_j} {x}
_{k}
\end{equation}
\end{small}
\vspace{-10pt}

\noindent where \(\vec{\mathbf{a}}\) is a single-layer feed-forward projection, and \(\bm{W}_{v}\), \(\bm{W}_{t}\) donated the transformation matrices, respectively.

\noindent\textbf{Hyperedge Weighting and Aggregation.}
The hypergraph topology is represented by an incidence matrix $\bm{H} \in \mathbb{R}^{|\mathcal{V}| \times |\mathcal{E}|}$. Each element $\mathbf{h}_{ij}$ is an attention coefficient quantifying the importance of hyperedge $\bm{e}_{j}$ to a given node $\bm{x}_{i}$:

\vspace{-7pt}
\begin{small}
\begin{equation}
    \mathbf{h}_{ij} = \frac{\exp \left( \text{LeakyReLU} \left( \vec{\mathbf{a}}^T [\bm{W} \bm{x}_{i} || \bm{W} \bm{e}_{j}] \right) \right)}{\sum_{k \in \mathcal{N}(i)} \exp \left( \text{LeakyReLU} \left( \vec{\mathbf{a}}^T [\bm{W} \bm{x}_{i}|| \bm{W} \bm{e}_{k}] \right) \right)}
\end{equation}
\end{small}

\vspace{-2pt}
\noindent where \(\mathcal{N}(i)\) denotes the neighborhood set of the \(i\)-th node. 

\noindent\textbf{Message Update and Iteration.}
Inter-node communication proceeds via hypergraph convolution. Mathematically,
\begin{equation}
\bm{X}^{(l+1)} = \phi \left( \bm{D}_{v}^{-1/2} \bm{H} \bm{W}_{e} \bm{D}_{e}^{-1} \bm{H}^T \bm{D}_{v}^{-1/2} \bm{X}^{(l)} \bm{\Theta}_{g}^{l} \right)
\end{equation}

\noindent where $\bm{D}_{e}$ and $\bm{D}_{v}$ are the diagonal matrices of edge and vertex degrees. $\bm{\Theta}_{g}^{l}$ is a learnable parameter at layer $l$, $\bm{W}_{e}$ is the diagonal hyperedge weight matrix, and $\phi$ is the activation function. After processing through the hypergraph network, the output features $\bm{\tilde{X}}$ are split into visual node features \(\{\tilde{\mathbf{z}}_v^{i}\}_{i=1}^{N}\) and textual node features \(\{\tilde{\mathbf{x}}_t^{j}\}_{j=1}^{L}\) for decoding.

\noindent\textbf{Feature Decoding and Grounding.}
The updated visual and textual node features are processed by a multi-layer dynamic (MLD) attention to yield the probability distribution $\bm{P(Y|\tilde{X})}$ over the visual nodes for the final grounding $\bm{Y}$.

\vspace{-20pt}
\begin{small}
\begin{equation}
\bm{P(Y|\tilde{X})} = \phi_{\text{MLD}} [ \underbrace{\tilde{z}_v^1,\ \tilde{z}_v^2,\ \ldots,\ \tilde{z}_v^{N}}_{\text{visual nodes}} ,\overbrace{\tilde{x}_t^{1},\ \tilde{x}_t^{2},\ \ldots,\ \tilde{x}_t^{L}}^{\text{textual nodes}}]
\end{equation}
\end{small}
\vspace{-10pt}

\noindent where the $\tilde{x}^i_v$ and $\tilde{x}^j_t$ are the output node features from $\bm{\tilde{X}}$.

\subsection{Training}
\vspace{-3pt}
We adopt a two-stage training pipeline: (1) World-Model Rollout Pretraining $L_{rol}$, which supervises the latent dynamics to predict future scene evolution; 
and (2) Grounding Decision Supervision $L_{gro}$, which trains the grounding module for target localization. See \textbf{Appendix~\ref{Training}} for more details.

\begin{table}[t]
      \setlength{\abovecaptionskip}{0pt}
      \setlength{\tabcolsep}{0.015\linewidth}
      \begin{center}
      \resizebox{0.47\textwidth}{!}{
      \begin{tabular}{c|c|c|c|c|c}
          \bottomrule
          Dataset & Method  & Venue & val/val-g & test A/val-u  & test B/test-u\\
        \midrule
           & TransVG \cite{deng2021transvg} & {{ICCV}} & 81.02 & 82.72 & 78.35 \\
          & VILLA \cite{gan2020large} & {{NeurIPS}} & 81.65 & \underline{87.40} & 74.48 \\
          & VLTVG  \cite{yang2022improving} & {{CVPR}} & 83.21 & 86.78 & 78.45\\
          \makecell[c]{RefCOCO}& SeqTR  \cite{chen2024mpcct} & {{ECCV}} & 78.22 & 81.47 & 73.80 \\
          & TransCP  \cite{10342826} &{{TPAMI}} & \underline{84.25} & 87.38 & \underline{79.78} \\
         & \textbf{ThinkDeeper\textsuperscript{s}}   & -  &75.72{\color{red}}& 79.49 {\color{red}}  &77.22\\
         & \textbf{ThinkDeeper}  & -  &\textbf{85.74}~{\color{red}$\pm$0.4}& \textbf{87.78}~{\color{red}$\pm$0.5}  &\textbf{80.64}~{\color{red}$\pm$0.2}\\
         
          \midrule\midrule
          
           & FAOA \cite{yang2019fast} & {{ICCV}} & 56.81 & 60.23 & 49.6 \\
          & VLTVG  \cite{yang2022improving} & {{CVPR}} & 72.36 & 77.21 & 64.8 \\
          & MPCCT \cite{chen2024mpcct} &{{PR}} & 73.28 & 78.96 & 63.59 \\
          \makecell[c]{RefCOCO+} & TransCP  \cite{10342826} &{{TPAMI}} & 73.07 & 78.05 & 63.35 \\
          & VILLA \cite{gan2020large} & {{NeurIPS}} & \underline{76.05} & \underline{81.62} & \underline{65.70} \\
          & \textbf{ThinkDeeper\textsuperscript{s}}  & -  &70.58{\color{red}}& 73.07 {\color{red}}  &59.53\\
        & \textbf{ThinkDeeper}  &-  &\textbf{77.72}{\color{red}~$\pm$0.5}& \textbf{82.10}{\color{red}~$\pm$0.3}  &\textbf{66.71}~{\color{red}$\pm$0.4}\\
        
          \midrule\midrule
           & NMTree \cite{liu2019learning} & {{CVPR}}  & 64.62 & 65.87 & 66.44\\
          & VLTVG  \cite{yang2022improving} & {{CVPR}} & \underline{72.53} & 74.90 & 73.88 \\
          & RvG-Tree \cite{hong2019learning} & {{TPAMI}}  & - & 66.95 & 66.51 \\
          \makecell[c]{RefCOCO-g} & TransCP  \cite{10342826} &{{TPAMI}} & 73.07 & 78.05 & 63.35 \\
          & ReSC-Large \cite{yang2020improving} & {{ECCV}} & 63.12 & 67.30 & 67.20 \\
          & VILLA \cite{gan2020large} & {{NeurIPS}} & - & \underline{75.90} & \underline{75.93} \\
          & \textbf{ThinkDeeper\textsuperscript{s}}  & -  &65.42{\color{red}}& 69.97 {\color{red}}  &63.72\\
        & \textbf{ThinkDeeper}  &-  &\textbf{72.73}{\color{red}~$\pm$0.2}& \textbf{80.90}{\color{red}~$\pm$0.5}  &\textbf{77.72}~{\color{red}$\pm$0.4}\\
          \bottomrule
      \end{tabular}
      }
      \end{center}
      \vspace{-5pt}
      \caption{Quantitative comparison of ThinkDeeper and SOTA baselines using IoU\(_{0.5}\) metric. \textbf{Bold} and \underline{underlined} denote the best and second-best scores. ThinkDeeper\(^s\) is the model trained on 75\% of the training data. Results are averaged over $\ge 3$ random seeds.}
      \label{tab:sota}

  \end{table}

\section{Experiments}\label{Experiments}

\subsection{Experimental Setup}
\textbf{Data Segmentation.} 
We evaluate ThinkDeeper on three real-world datasets: Talk2Car, DrivePilot, and MoCAD \cite{LIAO2024100116}, collectively forming the Full test set. To probe robustness, we further segment DrivePilot and MoCAD into two specialized test sets: Long-text and Corner-case. The Long-text set includes commands exceeding 23 words, which we expanded up to 50 words using Qwen2-VL to test complex linguistic handling. Correspondingly, the Corner-case set comprises three subsets: (1) 165 visual constraint scenarios (e.g., occlusions, low visibility), (2) 175 multi-agent interaction scenarios, and (3) 185 ambiguous command scenarios, designed to evaluate robustness in challenging real-world conditions. Additionally, we benchmark performance on the standard RefCOCO/+/g datasets using the segmentation methodology from the classic VG model VLTVG \cite{yang2022improving}.

\noindent\textbf{Evaluation Metric.} In accordance with the C4AV challenge, we utilize the ${IoU}_{0.5}$ score as the evaluation metric.

\noindent\textbf{Implementation Details.}
ThinkDeeper is trained on 4 $\times$ NVIDIA A100 GPUs, with a two-stage training process: 15 epochs in the first stage and 40 epochs in the second. We use a batch size of 32 and the AdamW optimizer with an initial learning rate of $10^{-4}$. See \textbf{Appendix~\ref{Setups}} for more details.

\subsection{Comparison to State-of-the-arts (SOTA)} 
\textbf{Overall Performance.} Table \ref{VG} reports the performance of ThinkDeeper against SOTA baselines. ThinkDeeper consistently outperforms all SOTA baselines across all datasets. For Talk2Car and DrivePilot, our model achieves improvements of 7.9\%  and 2.7\% in ${IoU}_{0.5}$, respectively, over the best-performing baselines UNINEXT and CAVG. For the MoCAD dataset, it reduces errors by at least 3.8\%, indicating the superior generalization of world model-based design.

\noindent\textbf{Model Robustness.}
ThinkDeeper excels in challenging scenarios, surpassing the next best baseline (UNINEXT) by 7.2\% on the Corner-case set and 11.8\% on the Long-text set. Specifically, on the Long-text set, it achieves 74.08 ${IoU}_{0.5}$ (+5.28 over VLTVG), highlighting its robustness in parsing extended commands, resolving linguistic ambiguity, and grounding in multi-agent scenes. Conversely, we observe that recent LLMs (MiniGPT-v2, LLaVA-NeXT, Qwen2-VL) perform poorly on these grounding tasks, with scores lagging 15-25 points behind SOTA. This is expected, as these general-purpose models lack the inductive biases for high-precision localization. Their inability to exploit 3D spatial context and depth cues to resolve ambiguity underscores the need for our depth-aware and spatial-aware world-model design tailored to real-world autonomous driving development.

\begin{table}[t]
  \centering
    \centering
    \setlength{\tabcolsep}{3mm}
    \resizebox{0.9\linewidth}{!}{
        \begin{tabular}{ccccccc}
            \bottomrule
            \multirow{2}[4]{*}{Components} & \multicolumn{6}{c}{Ablation Methods} \\
            \cmidrule{2-6}          & A     & B     & C     & D     & E   \\
            \hline
            Vision Encoder & \ding{56} & \ding{52} & \ding{52} & \ding{52} & \ding{52}   \\
              SA-WM (-Future)  & \ding{52} & \ding{56} & \ding{52}& \ding{52}  & \ding{52}   \\
             SA-WM  & \ding{52} & \ding{52} &  \ding{56} &\ding{52} & \ding{52}   \\
            Cross-modal Decoder  & \ding{52} & \ding{52} & \ding{52} & \ding{56} & \ding{52}   \\
 \hline
 \rowcolor{blue!7}  \multicolumn{1}{c}{$IoU_{0.5}$}  & 72.33 & 68.27  & 62.70    & 70.42   & 77.27    \\
            \toprule
        \end{tabular}}%
        \vspace{-4pt}
  \caption{\begin{small}Ablation studies for core component in DrivePilot.\end{small}}

  \label{ablation}%
\end{table}%


\begin{figure}[htb]
  \centering
  \includegraphics[width=0.97\linewidth]{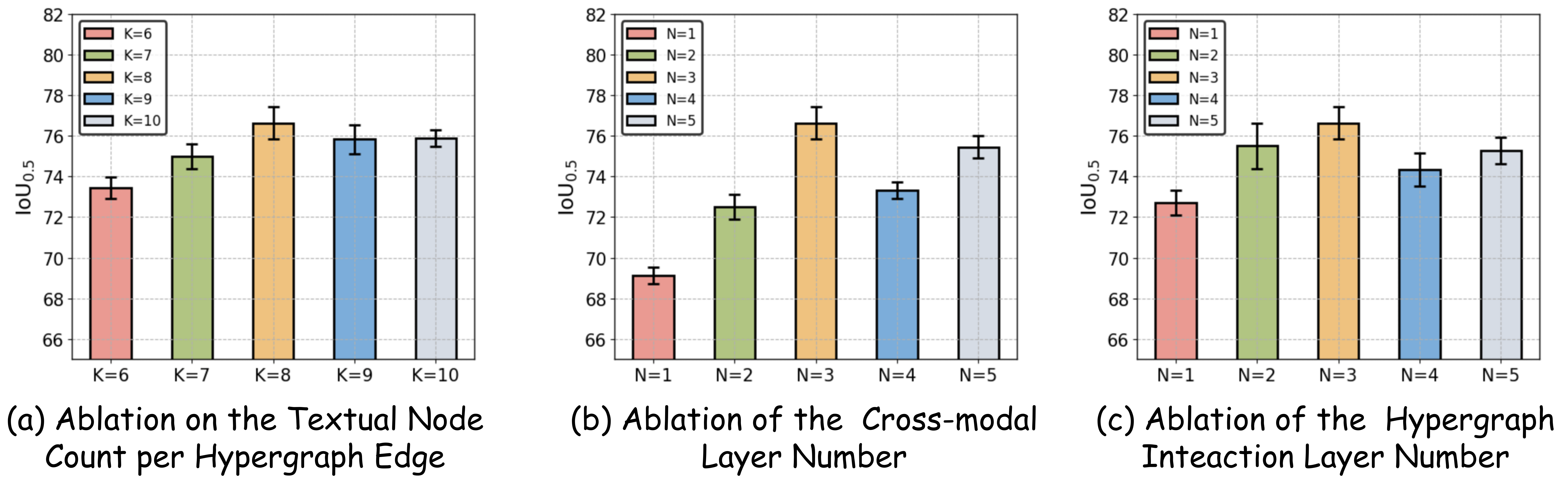}
    \vspace{-7pt}
  \caption{Ablation results of ThinkDeeper's hyperparameter.}
  \label{exp}
\end{figure}

 \begin{table}[t]
    \setlength{\abovecaptionskip}{0pt}
    \setlength{\tabcolsep}{0.015\linewidth}
    \begin{center}
    \resizebox{0.44\textwidth}{!}{
    \begin{tabular}{c|c|c|c|c}
        \bottomrule
         Method  & Backbone & Param. & Inference Time  & $IoU_{0.5}$ \\
        \hline
        VLTVG \cite{yang2022improving} & ResNet-101 & 152.18M & 55ms & 69.72 \\
        UNITER \cite{chen2020uniter} & ResNet-101 & 112.00M & 58ms & 73.42 \\
        RSDLayerAttn \cite{chan2022grounding} & ResNet-101 & 112.28M & 54ms & 74.12 \\
        CAVG \cite{LIAO2024100116} & ViT & 172.78M & 69ms & 74.50 \\
           \hline
        \rowcolor{blue!8}  
        \textbf{ThinkDeeper} & ViT & 135.81M & \textbf{39ms} & \textbf{78.93} \\
        \bottomrule
    \end{tabular}
    }
    \end{center}
   \vspace{-4pt}
    \caption{Efficiency comparison of ThinkDeeper and  baselines on Talk2Car, benchmarked using an NVIDIA A40 (40GB) GPU.}
    \label{tab:efficiency}
\end{table}

\begin{figure*}[htbp]
  \centering
  \includegraphics[width=1\linewidth]{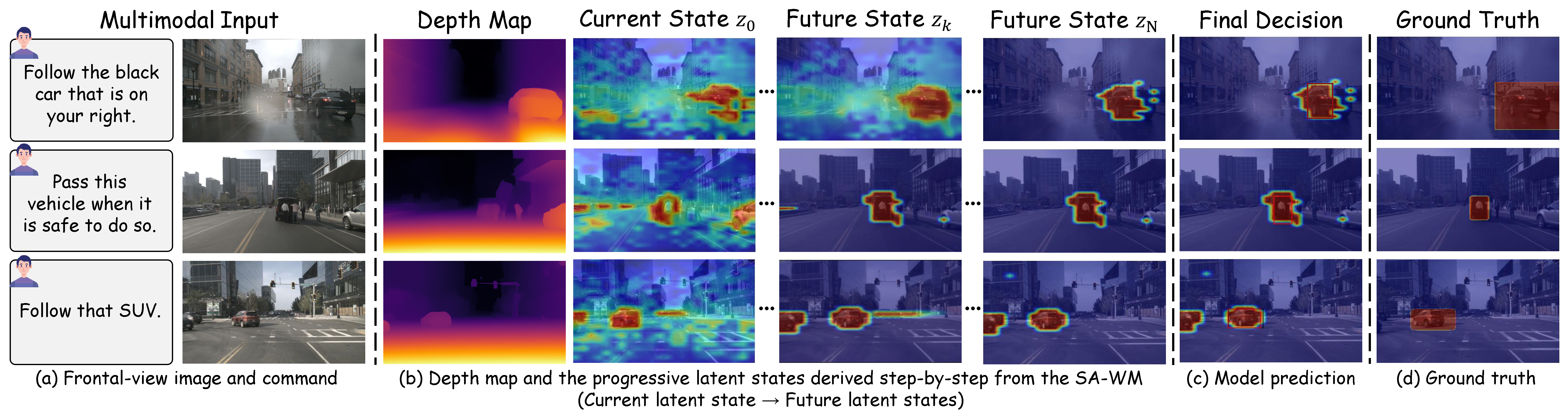}
  \vspace{-4mm}
  \caption{Qualitative results of depth map, current/future latent states, and model performance on the DrivePilot dataset.}
  \label{fig4}
\end{figure*}

\noindent\textbf{Model Generalizability.} We further evaluate ThinkDeeper on RefCOCO/+/g benchmarks. 
As reported in Table \ref{tab:sota}, it achieves impressive performance across all nine evaluation splits, surpassing the SOTA baselines by at least 2.9\%, 3.0\%, and 3.5\%, respectively. These results illustrate that although ThinkDeeper is developed for noisy, safety-critical on-road scenes, prioritizing lightweight real-time operation, it generalizes strongly to high-quality, densely annotated benchmarks, showcasing robust cross-domain applicability.

\noindent\textbf{Data Efficiency.} When trained on only 50\% or 75\% of the dataset, ThinkDeeper surprisingly beat most baselines trained on full datasets across all test sets. This shows its efficiency and scalability, reducing training data requirements while maintaining high performance in corner-case scenes.

\subsection{Comparison of Model Complex and Efficiency}
Table \ref{tab:efficiency} reports the inference efficiency of ThinkDeeper across 2,048 randomly selected scenes from Talk2Car, evaluated on an NVIDIA A40 GPU (70 TOPS). Despite utilizing fewer parameters than CAVG, our model still achieves the highest \(IoU_{0.5}\) (78.93), while maintaining a competitive average inference time of 39ms per sample, outperforming most SOTA baselines in both accuracy and efficiency. This performance meets the computational requirements for Level 3 AD systems (20-30 TOPS), enabling seamless deployment on various on-board Neural Processing Units (NPU) and Data Processing Units (DPU), such as Tesla FSD (245 TOPS) and NVIDIA Thor-U (500 TOPS).

\subsection{Ablation Study}
Table \ref{ablation} showcases the ablation study for each core component in ThinkDeeper. 
Our full model (Method E) achieves the highest score (77.27 $IoU_{0.5}$) and serves as the reference.
Specifically, Method A removes depth-derived prior from the Vision Encoder, causing a 6.3\% $IoU_{0.5}$ drop, underscoring the significance of depth information in refining spatial awareness. Method B omits the Future States Rollout, restricting the model to only the current latent state $z_0$. This incurs a severe 11.7\% performance drop, validating that static reasoning is insufficient and forward-looking inference is essential for resolving spatial ambiguity.  Method C removes the entire SA-WM module, revealing a catastrophic 14.57 $IoU_{0.5}$ collapse. This confirms that our world model's ability to distill a command-aware current latent state ($z_0$) and project future latent states ($\{z_k\}_{k=1}^N$) is a fundamental intermediate cue for robust grounding decisions.
Finally, Method D replaces our hypergraph with a standard GCN, which also suffers notable performance degradation. This highlights the hypergraph’s superior ability to encode visual–textual and multi-agent relations compared with pairwise graph updates.

\subsection{Hyperparameter Sensitivity Analysis}
Figure \ref{exp} reports our sensitivity analysis on the Talk2Car. We find that performance steadily improves as we increase the number of cross-modal attention layers, enabling richer interaction between the textual command and the SA-WM's latent states. The gains saturate at $N=3$, achieving our peak score of 76.64. Similarly, performance peaks when each visual node's hyperedge connects to $K=6$ textual nodes, indicating this provides an optimal balance of contextual information without introducing noise from irrelevant phrases. These results suggest that a moderately deep (3-layer) and broad (6-node) decoder is ideal for reasoning over our proposed world model's envisioned future states, while overly complex architectures yield diminishing returns.

\subsection{Visualization and Interpretability Analysis}
Figure \ref{fig4} reveals how our model ``thinks deeper'' by visualizing the latent states generated by the SA-WM in challenging scenes.  These states serve as intermediate evidence for the final grounding decision.  The SA-WM first distills the scene into a command-aware current latent state $z_0$, which encodes relevant objects, geometry, and saliency. It then rolls out future latent states $\{z_{k}\}_{k=1}^{N}$ that anticipate prospective changes in saliency and interaction patterns. This distilled, forward-looking representation provides the Decoder with a set of high-quality, filtered cues, enabling it to resolve ambiguity.
 For example, in the third row of Figure \ref{fig4}(b) (command: ``Follow that SUV.''), the model's initial attention is diffuse. However, as the SA-WM reasons about the ``follow'' and ``SUV'' concepts over its future states, it successfully isolates the target vehicle from background clutter. This internal filtering process directly enables the decoder to lock onto the correct SUV. These results support our quantitative findings: by treating current and imagined future latents as structured intermediate evidence, ThinkDeeper achieves robust grounding in low-light, occluded, crowded, and ambiguous scenes common in on-road applications. 
 We have included additional quantitative results of ThinkDeeper in \textbf{Appendix~\ref{Evaluation}}.

 \section{Conclusion}\label{Conclusion}
This paper introduces ThinkDeeper, the first world model-based framework for visual grounding in AD. By leveraging imagined future states as intermediate reasoning steps, ThinkDeeper effectively bridges the visual-linguistic gap, enabling more accurate and efficient grounding in real-world scenes. Importantly, we present DrivePilot, a multi-source dataset with extensive LLM-generated annotations, providing a challenging benchmark for VG research. ThinkDeeper achieves SOTA performance, ranking \#1 on the C4AV leaderboard for Talk2Car, DrivePilot, and MoCAD, while also showcasing impressive performance on the RefCOCO/+/g datasets, highlighting the potential of world models for robust cross-modal reasoning in fully autonomous driving.

\section*{Acknowledgements}
{\sloppy
This work was supported by the Science and Technology Development Fund of Macau
[0007/2025/RIC, 0122/2024/RIB2, 0215/2024/AGJ, 0074/2025/AMJ,
001/2024/SKL, 0002/2025/EQP], the Research Services and Knowledge Transfer Office,
University of Macau [SRG2023-00037-IOTSC, MYRG-GRG2024-\allowbreak00284-IOTSC],
the Shenzhen-Hong Kong-Macau Science and Technology Program Category C
[SGDX202308210951\allowbreak59012], the Science and Technology Planning Project
of Guangdong [2025A0505010016], National Natural Science Foundation of China
[52572354], the State Key Lab of Intelligent Transportation System [2024-B001],
and the Jiangsu Provincial Science and Technology Program [BZ2024055].
\par}

{
    \small
    \bibliographystyle{ieeenat_fullname}
    \bibliography{main}
}

\clearpage
\appendix
\setcounter{page}{1}


\appendix
\setcounter{page}{1}
\renewcommand{\thefigure}{A\arabic{figure}}
\renewcommand{\thetable}{A\arabic{table}}
\setcounter{figure}{0}
\setcounter{table}{0}
\setcounter{section}{0}

{
\centering
\Large
\textbf{Appendix}\\
}

\section{DrivePilot Dataset}\label{Dataset_process}
\subsection{Step-1: In-Context RAG Annotation}
To enhance LLM reasoning with real-world driving knowledge, we implement a two-tier Retrieval-Augmented Generation (RAG) framework. This process grounds AV visual grounding annotations in empirical driving data, ensuring contextual accuracy and reduced hallucination rates. We curate a multimodal knowledge base, including comprehensive expert-annotated scenarios from the nuScene dataset, covering agent trajectories, road topology, and traffic rule compliance. For each query pair (input image and command), we execute a three-phase retrieval process:

\noindent\textbf{Feature Encoding.} A pre-trained vision backbone (Fast R-CNN) extracts dense visual embeddings from the input image, capturing spatial relationships and object semantics. Simultaneously, a BERT-based language model encodes textual features from the given command, ensuring a rich semantic representation for cross-modal alignment.

\noindent\textbf{Cross-Modal Retrieval.} The extracted embeddings are used to retrieve the top-$k$ ($k=5$) most relevant scenes from the knowledge base via cosine similarity. The retrieved samples include both human-annotated metadata and raw sensor data from the traffic reports and nuScene dataset, enhancing scene understanding and context recall.

\noindent\textbf{Knowledge Infusion.} 
We template the retrieved scenarios into a structured prompt, instructing the LLM to ground its reasoning in both common driving behaviors, such as yielding to pedestrians in rain, and traffic regulations like interpreting ambiguous signals. This retrieval-augmented prompting ensures that the model's decisions are contextually grounded, enhancing both situational awareness and the trustworthiness of its reasoning.

\begin{figure}[t]
  \centering
  \includegraphics[width=\linewidth]{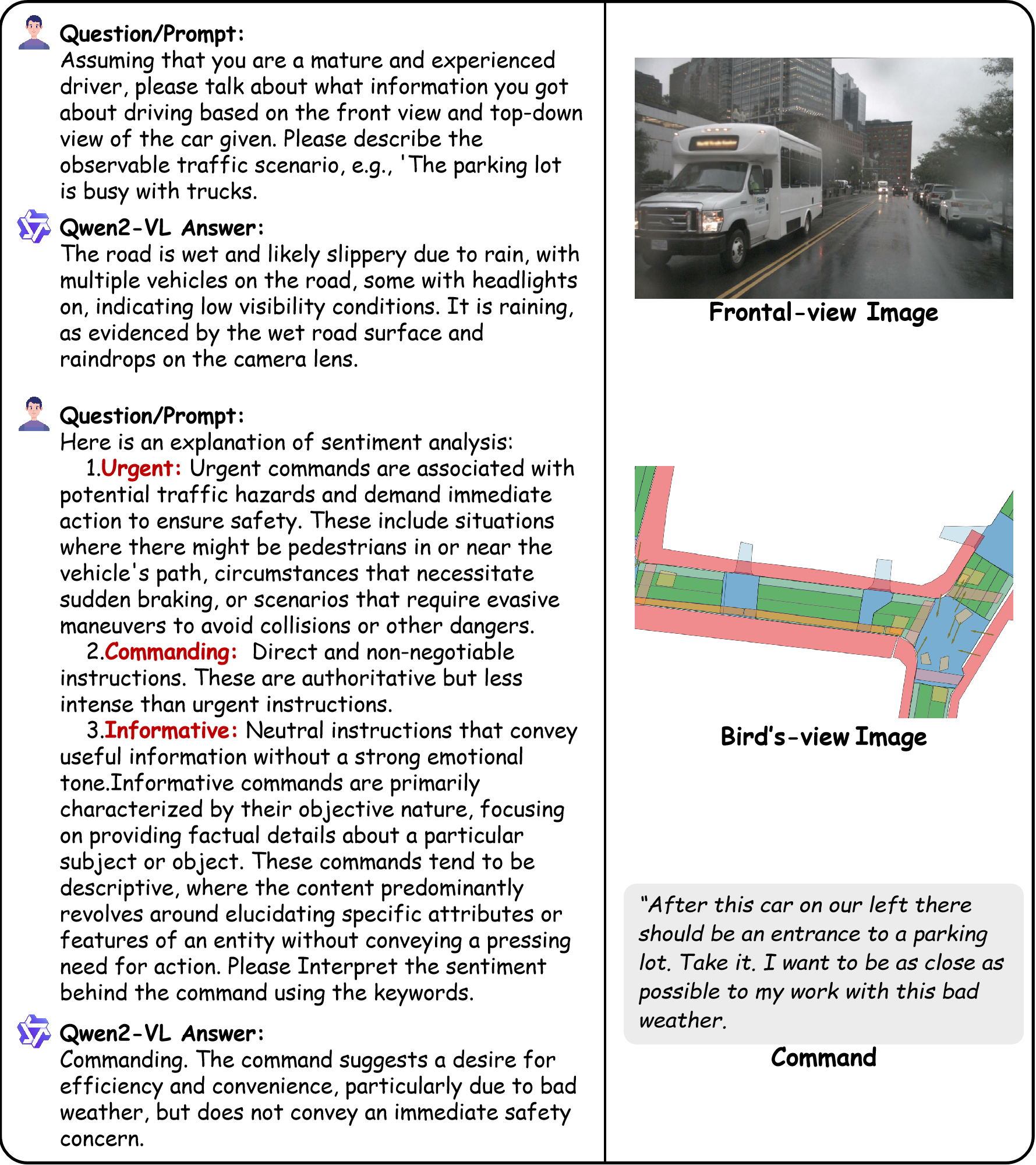}
  \caption{An example of the Qwen2-VL leveraging lens-less cueing technology to interpret driving scenarios and generate driving maneuvers, eliminating the need for specialized fine-tuning.}
  \label{PE5}
\end{figure}

\begin{table*}[htbp]
\centering
\begin{tabular}{>{\raggedright\arraybackslash}m{4cm} m{12cm}}
 \toprule
\textbf{Image Annotations }  \\ 
\midrule
\textbf{Visual input} & \begin{figure}[H]
  \begin{minipage}{1.4\linewidth}
    \centering
    \includegraphics[width=\linewidth]{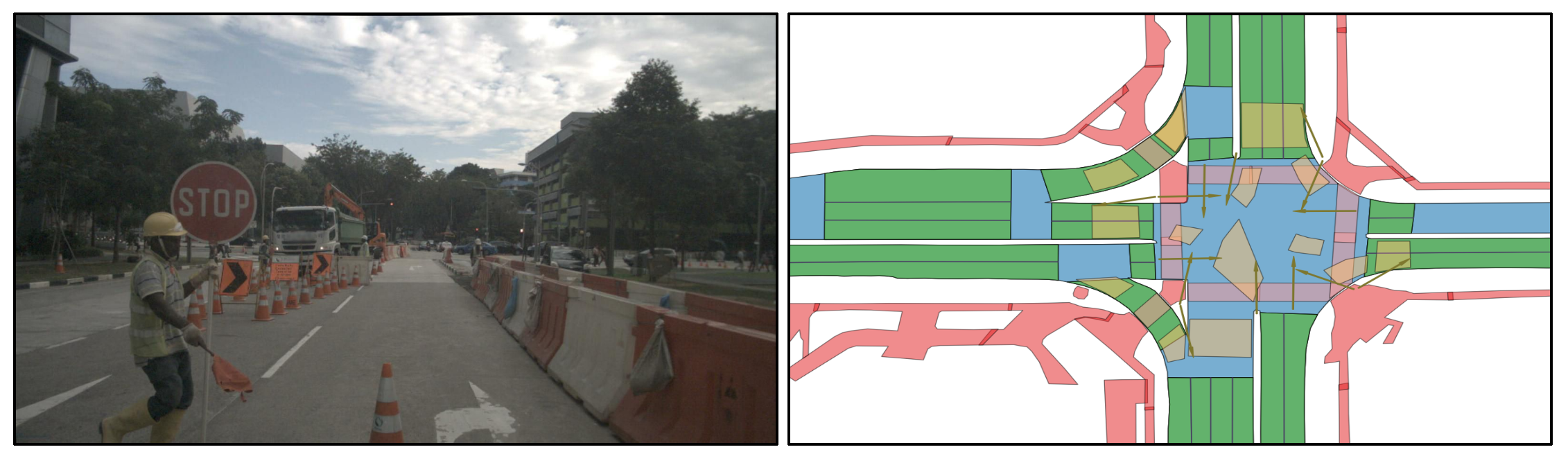}
    \captionsetup{width=\linewidth}
    \caption{An example of how a large language model interprets visual information: the left image represents the front view, while the right image corresponds to the BEV image.}
    \label{fig:extreme-ironing}
  \end{minipage}
\end{figure} \\
\midrule
 \textbf{Prompt} &
You are an AI visual assistant tasked with providing semantic enhancement for visual grounding tasks. Given an image containing multiple bounding boxes and its corresponding BEV (Bird's-Eye View) map, your role is to annotate each region by supplementing and describing the objects within it. In the BEV map, green areas represent drivable regions, while red areas indicate non-drivable regions. You should concisely summarize the behavior of objects in each region and their interactions with the surrounding environment, then describe the specific content of the provided bounding boxes, including object types, actions, positions, and relative locations between objects. Below are the coordinates for each entity object in the forward view and BEV map, along with the corresponding annotation. Example:
\begin{itemize}
    \item \textbf{Entities Annotation 1:} (label 1: [x1,y1,x2,y2]) Description of the features in this scene region.
    \item \textbf{BEV Annotation 1:} (label 1: [x1,y1,x2,y2]) Description of the spatial positioning in this region.
\end{itemize}
 \\
  \textbf{Entities Bounding box} &Person:[0.116,0.45,0.314,0.998], Truck[0.322,0.403,0.488,0.609],\\& Sign: [0.202,0.338,0.329,0.553], Car [0.685,0.541,0.756,0.630],  etc.\\
   \textbf{BEV coordinates}  & Entities BEV coordinates, Drivable area coordinates, Non-drivable area coordinates.\\
\midrule
 \textbf{LLM Annotations} & \begin{minipage}[t]{12cm}
  \textbf{Entities Annotation 1:} (Person: [0.116,0.45,0.314,0.998]) A worker in safety gear is holding a stop sign, managing traffic at a road construction site.\\[1ex]
  \textbf{Entities Annotation 2:} (Truck: [0.202,0.338,0.329,0.553]) A truck is involved in construction activities or waiting to proceed, on the left side of the road ahead.\\[1ex]
  \textbf{Entities Annotation 3:} (Stop Sign: [0.202,0.338,0.329,0.553]) A warning sign being held by a worker crossing the road.\\[1ex]
    \textbf{Entities Annotation 4:} (Car: [0.685,0.541,0.756,0.630]) A black sedan approaching head-on in the right lane.\\
  \textit{...}\\[1ex]
      \textbf{BEV Annotation 4:} (Car: [0.685,0.541,0.756,0.630]) 
A sedan positioned at the intersection behind a truck.\\[1ex]
\end{minipage} \\
  \toprule
\end{tabular}
\caption{Illustration of LLM-driven scene understanding. Given paired front-view images and BEV maps, the LLM is prompted with region definitions, bounding boxes, and BEV coordinates to produce rich semantic annotations. The prompt instructs the model to summarize scene context, describe object behaviors, and provide detailed entity-level grounding (types, actions, positions, and relations). By integrating 2D visual data with 3D spatial coordinates (BEV), the model generates fine-grained semantic descriptions. This demonstrates the LLM's ability to reason about object behaviors and interactions within a dynamic traffic environment.}
\label{tab:example}
\end{table*}

\subsection{Step-2: CoT-based Semantic Annotation}

LLMs like Qwen and LLaVA excel in natural language understanding but are not inherently trained for AD or VG tasks. Prior studies \cite{maniparambil2023enhancing, espejel2023gpt,liao2025cot} have shown that structured prompt engineering can significantly improve LLMs' zero-shot visual description performance. To leverage this potential, we develop a progressive Chain-of-Thought (CoT) prompting strategy for generating context-aware semantic annotations, enabling dataset augmentation and semantic enrichment without fine-tuning. Specifically, we explore the utility of Qwen in augmenting and refining existing multimodal data (frontal and BEV images, paired with natural language commands) using few-shot or zero-shot prompting techniques. This enables dataset expansion and improved semantic annotation generation without costly and time-consuming fine-tuning. As illustrated in Table \ref{tab:example} and Figure \ref{PE5}, our CoT framework guides the model (Qwen2-VL) through $h$ reasoning iterations to extract 14 distinct semantic dimensions. This CoT process includes:
\begin{itemize}
    \item Scene Descriptions: Summarizing overall environmental context, including road surface conditions, weather, lighting, and potential hazards in the traffic scenes.

    \item Emotion Interpretation: Analyzing pedestrian and driver intent to predict potential interaction risks.

    \item Road Condition Summaries: Detailing surface quality, lane markings, and obstacles affecting navigation.

    \item Traffic Signal \& Sign Interpretations: Detailing surface quality, lane markings, and obstacles.

\end{itemize}

To diversify training data, we randomly replace 30\% of the original command texts with this CoT-generated augmentation text during training, ensuring diverse linguistic patterns and robust generalization. Furthermore,  keyword-based augmentation is introduced to append relevant context cues to the original command prompts, aiding semantic disambiguation. For example, semantic keywords such as ``low visibility'' and ``intersection'' are added as auxiliary hints to commands in scenarios with obstructed vision or Multi-agent. Notably, this step-by-step process involves dialogues where each ``thought'' guides Qwen2-VL to understand different aspects of the scene or command. The first thought focuses on understanding the scene and identifying key objects and their dynamics. The second thought analyzes command keywords and emotions. After \( h \) iterations of reasoning and updating (with iterations varying per sample for the actual situation), insights from each thought are synthesized into a cohesive semantic scene annotation.

\subsection{Step-3: Manual Cross-check Validation} 
To ensure high fidelity and compliance, all LLM-generated annotations undergo a rigorous multi-stage review by 13 domain experts, including AV safety engineers and certified instructors. Annotations are validated against nuScenes sensor data (LiDAR, radar, and camera) to ensure spatial precision. Specifically, object positions and spatial relationships are corroborated with 3D ground-truth coordinates, while temporal consistency is verified across frames to align with actual motion trajectories. Furthermore, BEV annotations are manually audited to resolve spatial ambiguities and ensure precise depth perception, eliminating false positives.

\section{Spatial-Aware World Model}\label{SA_WM}
The calculation of depth-derived prior \(P(k)\) is derived from the depth graph \(F_d\), which encodes the depth information of the input visual data. To ensure a consistent depth representation, \(F_d\) is normalized to the range \([0, 1]\) using an Exponential Decay Function. This function assigns higher values to closer objects while attenuating the influence of distant regions, ensuring depth-aware feature refinement. The transformation can be formally expressed as follows:
\begin{equation}
F_{D}^{\text{nor}}(x) = \exp(-\alpha \cdot F_{D}(x))
\end{equation}
where \(\alpha\) is a decay rate hyperparameter that regulates depth sensitivity, preserving finer details for nearby objects while suppressing distant regions. Pixels corresponding to objects at infinite depth are set to zero, excluding them from further processing to improve computational efficiency.

Next, the normalized depth map \(F_{D}^{\text{nor}}\) is passed through a Multi-Layer Perceptron (MLP), which employs a piecewise activation function. This allows the model to adaptively emphasize depth regions based on their visual importance, refining spatial awareness in visual grounding. Formally,
\begin{equation}
P(x) = \phi_{\text{MLP}}(F_{D}^{\text{nor}}(x))
\end{equation}
where \(\phi_{\text{MLP}}\) represents the MLP transformation, mapping depth-normalized features for downstream tasks.

\section{Training Loss}\label{Training}

To train our model, we define supervision over the prediction tuple $\{s, y\}$ and the ground-truth tuple $\{\hat{s}, \hat{y}\}$, where $Z_v$ denotes the visual latent states produced by SA-WM. For each visual block, the model is required to regress the IoU between the predicted region and the annotated ground-truth region, which is denoted as $y$. This design ensures that every visual block contributes explicit grounding signals rather than relying solely on a final aggregated score.

\noindent\textbf{Stage-1: World-Model Rollout Pretraining.}
In the first stage, the supervision is dominated by dense, patch-level observations in complex traffic scenes. Under such settings, standard region-based losses like Dice loss or vanilla cross-entropy frequently suffer from severe foreground–background imbalance, caused by large fields of view, multiple reference objects, and a relatively small proportion of positive pixels. To mitigate this, we draw inspiration from lesion detection in medical imaging and combine a Tversky loss $\mathcal{L}_{\mathrm{tve}}$ \cite{salehi2017tversky} with Focal loss $\mathcal{L}_{\mathrm{foc}}$ \cite{lin2017focal}.  During World-Model rollouts, we initialize the prior score $S$ with uniform probability to accelerate the convergence of visual-text mapping. By leveraging bilinear interpolation on the downsampled ground-truth mask $\hat{{S}}$, we enforce a dense alignment constraint.

Unlike standard formulations that simply sum losses, we construct a weighted integration of a re-parameterized Tversky objective and a hardness-weighted Focal penalty. Let $p_{ij} \in {S}$ and $g_{ij} \in \hat{{S}}$ denote the predicted probability and binary ground truth at spatial location $(i,j)$. We define the intersection and set differences for the $k$-th class as follows:
\begin{align}
    \mathcal{I}_k &= \sum\nolimits_{i,j} p_{ij} g_{ij}, \nonumber \\
    \mathcal{F}_k^{\alpha} &= \alpha \sum\nolimits_{i,j} (1-g_{ij}) p_{ij}, \;\;
    \mathcal{F}_k^{\beta} = \beta \sum\nolimits_{i,j} (1-p_{ij}) g_{ij}
\end{align}
\noindent where $\epsilon$ is a smoothing factor, and $\alpha, \beta$ control the penalty magnitude for False Positives (FP) and False Negatives (FN), respectively, allowing the model to dynamically shift focus during training. Then, the rollout  loss ${L}_{rol}$ is formulated as:

\vspace{-8pt}
\begin{small}
\begin{equation}
\label{eq:rollout_loss}
\begin{split}
{L}_{\text{roll}} = \lambda_{tve}\sum_{k=1}^{K} \left[ 1 - \frac{\mathcal{I}_k + \epsilon}{\mathcal{I}_k + \mathcal{F}_k^{\alpha} + \mathcal{F}_k^{\beta} + \epsilon} \right] 
 - \lambda_{\text{foc}} \frac{1}{N} \sum_{n=1}^{N} \mathcal{H}(p_n, g_n)
\end{split}
\end{equation}
\end{small}

Here, the $\lambda_{tve}$ and $\lambda_{foc}$ are both the hyperparameters. Moreover, $\mathcal{H}(\cdot)$ represents the focal modulation defined as 
$\mathcal{H}(p_n, g_n) = \alpha_t (1 - p_{t,n})^\gamma \log(p_{t,n})$, 
where $p_{t,n}$ reflects the model's confidence in the true class. This formulation ensures that gradients are dominated by hard-to-classify examples like small distant vehicles, rather than the dominant background, effectively regularizing the rollout trajectory.

\noindent\textbf{Stage-2: Grounding Decision Supervision.}
Upon establishing robust feature representations, the second stage focuses on precise localization. To prevent catastrophic forgetting of the patch-level priors learned in the first stage, we employ a multi-task learning strategy. We define the grounding loss ${L}_{\text{ground}}$ as a hybrid constraint optimization that simultaneously minimizes the distributional divergence and the geometric regression error. Let $\hat{\mathbf{y}}$ be the ground truth bounding box and ${S}_{lat}$ be the latent state output representing the object mask. The objective function aggregates the Binary Cross-Entropy (BCE) and L1 regression error, normalized over the batch. The supervision loss ${L}_{\text{gro}}$ can be defined as:
\begin{equation}
\label{eq:grounding_loss}
{L}_{\text{gro}} = \mathbb{E}_{({y}, {S}) \sim \mathcal{D}} \left[ \lambda_{\text{cls}} \cdot \Psi_{\text{BCE}}({y}, \hat{{y}}) + \lambda_{\text{reg}} \cdot \|{S}_{lat} - \hat{{S}}\|_1 \right]
\end{equation}

\noindent where $\Psi_{\text{BCE}}$ is the binary cross-entropy operator, and $\|\cdot\|_1$ imposes a sparsity-inducing L1 penalty on the mask prediction. The hyperparameters $\lambda_{\text{cls}}$ and $\lambda_{\text{reg}}$ balance the trade-off between semantic classification accuracy and geometric precision. Overall, this diversity loss term incentivizes the model to capture cross-modal interactions and produce target predictions consistent with the commander's intent.

\begin{figure*}[htbp]
  \centering
  \includegraphics[width=1\linewidth]{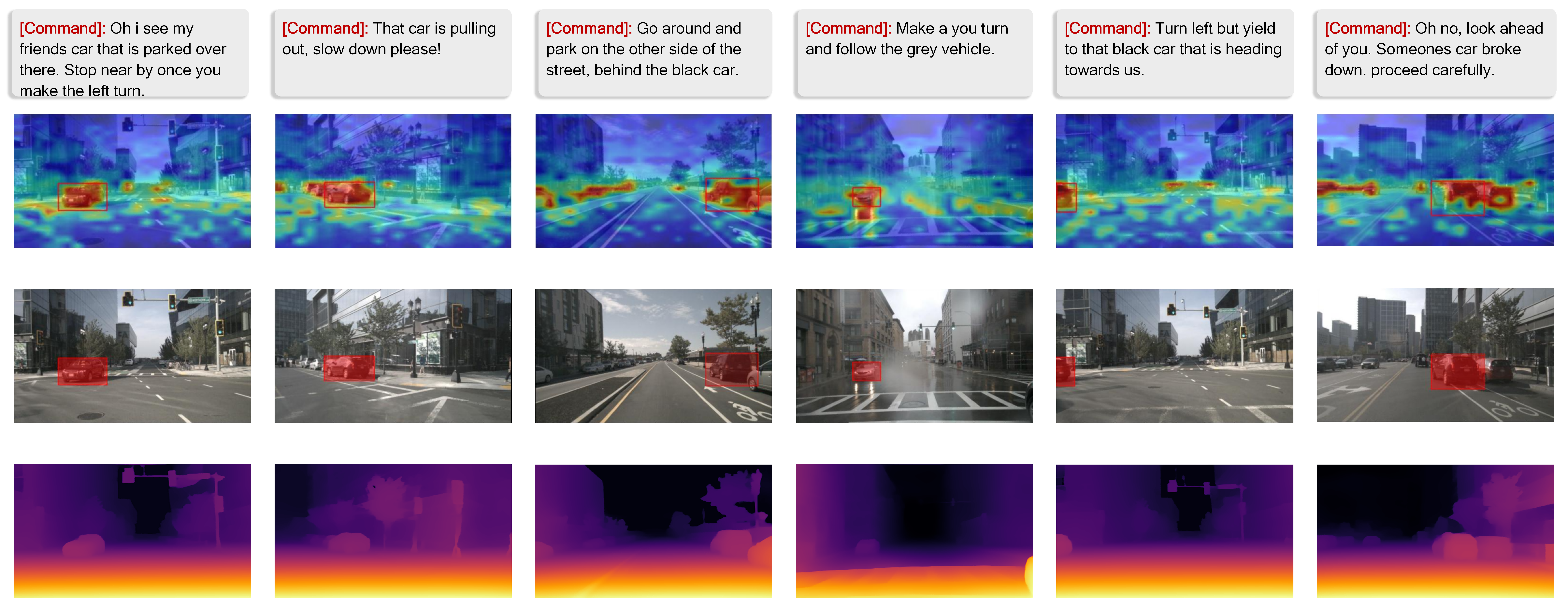}
  \caption{Visualization of ThinkDeeper's multimodal grounding. The first line corresponds to the query commands for each scene, while the second row shows the future latent states $Z_v$ generated by the SA-WM alongside the corresponding prediction. The third row highlights the ground truth regions in red mask areas, while the fourth row presents the depth map, providing spatial context for scene understanding.}
  \label{visualization}
\end{figure*}

\section{Experiments Setups}\label{Setups}
\subsection{Benchmarks}
To evaluate our model's effectiveness, we conduct experiments on the dataset zoo: Talk2Car, DrivePilot, MoCAD, and RefCOCO, RefCOCO+, and RefCOCOg. These datasets provide diverse and complex real-world scenarios for benchmarking visual grounding in the field of autonomous driving.

\noindent\textbf{Talk2Car.} The Talk2Car dataset \cite{deruyttere2019talk2car}, an extension of the NuScenes dataset \cite{caesar2020nuscenes}, consists of 11,959 natural language commands across 9,217 images captured in urban landscapes of Singapore and Boston. This dataset includes a variety of conditions, such as different times of day and weather scenarios, offering a challenging and diverse benchmark. The commands, averaging 11 words, contain complex instructions (e.g., ``Parallel park behind the black car on our right''), requiring precise semantic understanding and scene reasoning. It enhances the NuScenes with bounding box annotations across 850 videos, with 55.94\% of commands originating from Boston and 44.06\% from Singapore. A detailed linguistic analysis reveals an average of 11.01 words per command, comprising 2.32 nouns, 2.29 verbs, and 0.62 adjectives, highlighting the linguistic diversity and complexity of the dataset. Each video is associated with an average of 14.07 commands, enriching the contextual learning process.

\noindent\textbf{DrivePilot.} We introduce DrivePilot, the first dataset to leverage Qwen’s linguistic capabilities for detailed semantic scene annotation using regularized prompts. This dataset categorizes urban scenes across 14 dimensions, including weather conditions, emotional context, and agent interactions. Each dataset entry comprises a natural language command, paired front-view and BEV images, scene annotations generated by Qwen2-VL, and precise target object locations. The dataset is designed to challenge models with object disambiguation and complex query interpretation, closely reflecting real-world AV navigation challenges.

\noindent\textbf{MoCAD.} The dataset originates from the first Level 4 autonomous bus deployed in Macau and has been continuously tested since 2020. The dataset spans 300+ hours of real-world driving, including data sets from a 5-kilometer campus route, a more extensive 25-kilometer city and urban road collection, and various open traffic situations observed under varying weather, time, and traffic density conditions. It comprises over 13,000 scene images and nearly 40,000 scene objects, with an average command length of 12.5 words, providing a rich dataset for visual grounding research. A distinctive aspect of MoCAD is its Macau-based driving environment, where right-hand driving contrasts with regions that enforce left-hand driving. This contrast introduces unprecedented challenges for VG in AVs, particularly in terms of context adaptation and cross-domain generalization.

\noindent\textbf{RefCOCO.} RefCOCO is built from the ReferItGame, a two-player interface for collecting referring expressions. It contains 142,209 expressions for 50,000 objects in 19,994 images. The dataset is split into train/val/test sets, with the test set further divided into Test A (images with multiple people) and Test B (images with only objects).

\noindent\textbf{RefCOCO+.}
 Similar to RefCOCO, this dataset was also collected through the ReferItGame but with a restriction that prohibits the use of location-based descriptions. This constraint encourages the use of appearance-based expressions, making the task more challenging. RefCOCO+ comprises 141,564 referring expressions for 49,856 objects in 19,992 images. The dataset shares the same split structure as RefCOCO, including the Test A and Test B subdivisions.

\noindent\textbf{RefCOCOg.}
 Collected in a non-interactive setting, RefCOCOg features longer and more complex referring expressions, averaging 8.4 words per expression compared to 3.5 words in RefCOCO. It includes 95,010 expressions for 49,822 objects across 25,799 images. It is split into training, validation, and test sets, focusing on more descriptive and detailed language, which poses additional challenges for language comprehension and visual grounding models. 

\subsection{Implementation Details}
\noindent\textbf{Overall Configuration.} Input images are resized to 384×384 pixels, and text expressions are truncated to a maximum of 50 tokens. During training, we apply a text augmentation strategy: with 30\% probability, original descriptions are replaced by LLM-augmented versions, followed by keyword appending after a [SEP] token. We use the AdamW optimizer with a batch size of 32 and adopt a learning rate warmup over the first 10\% of training steps. All components except the vision-language extractors (ViT and BERT) are trained with an initial learning rate of $10^{-4}$. ViT and BERT are initialized using BLIP \cite{li2022blip} pre-trained weights, while other modules employ Xavier initialization \cite{glorot2010understanding}.

\noindent\textbf{Text Encoder.}  We use a pre-trained BERT model for text embedding extraction, configured with 16 hidden layers and an embedding vocabulary of 30,524 tokens. We also enforce a maximum sentence length of 50 tokens, with a layer normalization epsilon of 1e-12, and a hidden size calibrated to $d=768$ for linguistic input processing.

\noindent\textbf{Vision Encoder.} We use a Vision Transformer-Base (ViT-B) as the vision encoder with a 4:1 MLP to embed dimension ratio and 12-head multi-head attention, extracting a 24×24 visual token stream, and 3 cross-modal attention layers.

\noindent\textbf{Spatial-Aware World Model.} We use a three-layer cross-modal attention layer ($N=3$) to compute prior scores, where each layer’s output is projected through a linear head. Each cross-modal attention block uses a hidden size of \(D=768\) for both text and visual inputs, with 8 attention heads and a dropout rate of 0.1. The learnable parameters in the discriminative module are initialized with \(\mu = 1.0\) and \(\sigma = 1.0\). Training proceeds in two stages: we first optimize the model with the world-model rollout loss $L_{rol}$ for 15 epochs, followed by grounding-focused training with $L_{gro}$ for an additional 55 training epochs (i.e., 70 epochs total).

\noindent\textbf{Multimodal Decoder.}
Our architecture employs a cross-modal hypergraph where each visual node connects to its top \(L=8\) text nodes to form hyperedges, selected according to an affinity matrix computed by a 1536-dim MLP. Hypergraph attention uses 4 heads with LeakyReLU (negative slope 0.2), a 768-dim hidden layer, and 0.2 dropout. The multi-layer dynamic attention stack consists of 6 attention blocks (each followed by a linear layer), with 12-head multi-head attention, 768-dim hidden size, and 0.2 dropout.

\subsection{Corner-case and Long-text Test Sets}
To assess model robustness under real-world challenging conditions, we curate four specialized test subsets from the DrivePilot and MoCAD datasets. These include restricted visibility, multi-agent interactions, ambiguous prompts, and long, complex commands. In the multi-agent set, we focus on scenes containing more than 16 targets. The visual constraints set consists of scenarios with impaired visibility caused by nighttime conditions, fog, rain, camera obstructions, or low-resolution images. To evaluate the model’s ability to handle linguistic ambiguity, we identify unclear or potentially ambiguous commands, categorizing them as the ambiguous set. Additionally, recognizing that longer commands often introduce irrelevant details or increased complexity, we select commands exceeding 23 words and categorize them into the long-text test set, designed to assess the model’s capacity to process intricate instructions.

\section{Quantitative Results}\label{Evaluation}
As shown in Figure \ref{visualization}, we present more quantitative results generated by ThinkDeeper in the DrivePilot dataset. These qualitative results showcase the superior performance of our proposed world model-inspired framework in integrating multimodal, real-world commands by jointly leveraging language, spatial, and visual cues, leading to improved multimodal grounding accuracy. These examples highlight our model's ability to achieve robust localization under challenging conditions like high multi-agent and ambiguous scenes.


\end{document}